\documentclass{article}
\usepackage{ijcai20}

\usepackage{times}

\usepackage{soul}
\usepackage{url}
\usepackage[hidelinks]{hyperref}
\usepackage[utf8]{inputenc}
\usepackage[small]{caption}
\usepackage{graphicx}
\usepackage{amsmath}
\usepackage{booktabs}
\usepackage{amsmath}
\usepackage{amsfonts}
\usepackage{xcolor}
\usepackage{enumitem}
\usepackage{comment} 
\usepackage{graphicx}
\PassOptionsToPackage{numbers}{natbib}
\usepackage{listings}

\newcommand{\joey}[1]{\textcolor{red}{[JOEY: #1]}}

\usepackage{amsmath}
\usepackage{amsfonts}
\usepackage{xcolor}
\usepackage{comment} 
\usepackage{graphicx, subcaption}

\newcommand{\red}{\textcolor{red}}

\newcommand{\xhdr}[1]{{\noindent\bfseries #1}.}
\newcommand{\cut}[1]{}

\newcommand{\removelatexerror}{\let\@latex@error\@gobble}
\newcommand{\namelong}{GAALV}
\newcommand{\nameshort}{GAALV}
\newcommand{\aename}{GAALV-AE}

\title{Generalizable Adversarial Attacks with Latent Variable Perturbation Modeling}

\author{
Avishek Joey Bose $^{1,2}$
\and
Andre Cianflone $^{1,2}$\And
William L. Hamilton$^{1,2}$\
\affiliations
$^1$McGill University\\
$^2$Mila\\
\emails
\{joey.bose, andre.cianflone, wlh@cs\}.mcgill.ca
}

\begin{document}
\maketitle

\begin{abstract}
Adversarial attacks on deep neural networks traditionally rely on a constrained optimization paradigm, where an optimization procedure is used to obtain a single adversarial perturbation for a given input example. 
In this work we frame the problem as learning a distribution of adversarial perturbations, enabling us to generate diverse adversarial {\em distributions} given an unperturbed input.
We show that this framework is \textit{domain-agnostic} in that the same framework can be employed to attack different input domains with minimal modification. 
Across three diverse domains---images, text, and graphs---our approach generates whitebox attacks with success rates that are competitive with or superior to existing approaches, with a new state-of-the-art achieved in the graph domain.
Finally, we demonstrate that our framework can efficiently generate a diverse set of attacks for a single given input, and is even capable of attacking \textit{unseen} test instances in a zero-shot manner, exhibiting {\em attack generalization.}
\end{abstract}

\section{Introduction}\label{sec:intro}

Adversarial attacks on deep learning models involve adding small, often imperceptible, perturbations to input data with the goal of forcing the model to make certain  misclassifications 
\cite{szegedy2013intriguing}. 
\cut{
}There are a wide-variety of threat models that define settings and assumptions under which attack strategies are developed.
This includes the so-called ``whitebox'' setting, where the model parameters are accessible to the attacker, as well as the more challenging setting of ``blackbox'' attacks, where the attacker only has access to the final outputs of the target model \cite{papernot2016transferability,papernot2017practical}.

Despite this diversity, a commonality in most existing approaches is the treatment of generating adversarial attacks as a constrained optimization or search problem \cite{carlini2017towards}.
In this constrained optimization paradigm, the objective is to search for a specific adversarial perturbation based on a particular input datapoint, with the constraint that this perturbation is not too large. 
The constrained optimization paradigm has led to a number of highly successful attack strategies, with whitebox attacks based on first-order gradient information being particularly ubiquitous and successful.
Examples of successful strategies employing this approach include the classic Fast Gradient Sign Method (FGSM) \cite{goodfellow2014explaining}, as well as more recent variants such as L-BFGS \cite{szegedy2013intriguing}, Jacobian-based Saliency Map Attack (JSMA) \cite{papernot2016limitations}, DeepFool \cite{moosavi2016deepfool}, Carlini-Wagner \cite{carlini2017towards} and the PGD attack \cite{madry2017towards}, to name a few. 

Unfortunately, the constrained optimization paradigm has important limitations. 
For example, while constrained optimization-based attack strategies are easily applicable to continuous input domains (i.e., images), adapting them to new modalities, such as discrete textual data \cite{li2018textbugger,ebrahimi2017hotflip,gao2018black} or graph-structured data that is non-i.i.d. \cite{dai2018adversarial,zugner2018adversarial}, represents a significant challenge \cite{text_survey1,sun2018adversarial}. 
In addition, in the constrained optimization approach, a specific optimization step must be performed for each attacked input, which generally only leads to a single or a small set of non-diverse perturbations. 
As a consequence, constrained optimization approaches often fail to produce diverse attacks---which can make them easier to defend against---and these approaches do not efficiently generalize to unseen examples without further optimization.

\xhdr{Present work}
We propose \namelong\ ({\bf G}enerelizable {\bf A}dversarial {\bf A}ttacks with {\bf L}atent {\bf V}ariable perturbation modeling), a unified generative framework for whitebox adversarial attacks, which is easily adaptable to different input domains and can efficiently generate diverse adversarial perturbations. 
\namelong\ leverages advancements in deep generative modeling to learn an encoder-decoder based model with a stochastic latent variable that can be used to craft small perturbations. Our approach offers three key benefits: 
\begin{itemize}[leftmargin=*, topsep=0pt, itemsep=2pt, parsep=0pt]
    \item {\bf Domain-agnostic.} Our framework can be easily deployed in diverse domains (e.g., images, text, or graphs) by simply choosing an appropriate encoder, decoder, and similarity function for that domain. 
    \item {\bf Efficient generalization.}
    Employing a parametric model with a stochastic latent variable allows us to amortize the inference process when crafting adversarial examples.  After training, we can efficiently construct these adversarial examples and even generalize without any further optimization to unseen test examples with only a single pass through the trained network.
    \item {\bf Diverse attacks.} We learn a conditional distribution of adversarial examples, which can be efficiently sampled to produce diverse attacks or used to {\em resample} when an initial attack fails.
\end{itemize}
To demonstrate the benefits of \namelong\, we implement variants to attack CNN-based image classifiers,  LSTM-based text classifiers, and GNN-based node classification models. 
In the image domain, our framework achieves results that are competitive with constrained optimization-based approaches in terms of attack success rates, while providing more efficient generalization and diverse perturbations.
In the text domain, we provide strong results on the IMDB benchmark, being the first work to scale an approach with state-of-the-art success rates to the full test set. 
Finally, in the graph domain, our approach achieves a new state-of-the-art for attacking a GNN using node features alone.
\section{Background and Preliminaries}

Given a classifier $f : \mathcal{X}\rightarrow \mathcal{Y}$, input datapoint $x \in \mathcal{X}$, and class label $y \in \mathcal{Y}$, where $f(x) = y$, the goal of an adversarial attack is to produce a perturbed datapoint $x' \in \mathcal{X}$ such that $f(x') = y' \neq y$, and where the distance $\Delta(x, x')$ between the original and perturbed points is sufficiently small.

\xhdr{Threat models}
Adversarial attacks can be classified under different threat models,
which impose different access and resource restrictions on the attacker \cite{akhtar2018threat}. 
In this work we consider the whitebox setting, where the attacker has full access to the model parameters and outputs. This setting is more permissive than the blackbox \cite{papernot2016transferability,papernot2017practical} and semi-whitebox \cite{xiao2018generating} settings, which we consider for test set attacks. Constrained optimization attacks in the whitebox setting are relatively mature and well-understood, making it a natural setting to contrast our alternative generative modelling-based attack strategy.
In addition, we consider the more common setting of {\em untargeted} attacks, where the goal of the attacker is to change the original classification decision to any other class.\footnote{Our framework could be extended to the targeted setting by modifying the loss function, as well as to blackbox attacks via blackbox gradient estimation. We leave these extensions to future work.} 

\xhdr{Constrained optimization approach}
In the standard constrained optimization framework, the goal is to find some minimal perturbation $\delta \in \mathcal{X}$ that $\textnormal{minimizes} \; \Delta(x,x + \delta)$, subject to $ f(x + \delta) = y'$ and $x + \delta \in \mathcal{X}$,
\cut{
\begin{align}\label{eq:conopt}
& \textnormal{minimize} \; \Delta(x,x + \delta) \\
& \textnormal{ s.t. }  f(x + \delta) = y' \nonumber \\ 
& x + \delta \in \mathcal{X} \nonumber, 
\end{align}
}
where the last constraint is added to ensure that the resulting $x'=x+\delta$ is still a valid input (e.g., a valid image in normalized pixel space.) 
Rather than solving this non-convex optimization problem directly, the typical approach is to relax the constraints into the objective function and to employ standard gradient descent \cite{goodfellow2014explaining} or projected gradient descent, with the projection operator restricting $x'$ such that $\Delta(x,x') \leq \epsilon$ and $x' \in \mathcal{X}$ \cite{madry2017towards}.

\xhdr{Limitations of the constrained optimization approach}
The constrained optimization formulation \cut{(Equation \ref{eq:conopt})} has been effectively deployed in many recent works, especially for images (see \cite{akhtar2018threat} for a survey).
However, this approach has two key technical limitations:
\begin{enumerate}[leftmargin=15pt, topsep=5pt, itemsep=2pt, parsep=5pt]
    \item Directly searching for an adversarial example $x'$ or an adversarial perturbation $\delta \in \mathcal{X}$ can be computationally expensive if the input space $\mathcal{X}$ is very high-dimensional or discrete (e.g., for text or graphs). Moreover, adding the perturbation $\delta$ to $x$ is only feasible if the input domain $\mathcal{X}$ is a field with a well-defined notion of addition, which is not the case for discrete domains such as text. 
    \item A distinct optimization procedure must be run separately for each attacked datapoint, and without substantial modifications, this optimization will only yield a single (or small set) of non-diverse perturbations for each attacked datapoint. 
\end{enumerate}
Together, these limitations make it non-trivial to generalize the constrained optimization approach beyond the image domain, and make it difficult to generate diverse attacks for unseen datapoints. \cut{that generalize to previously unseen datapoints.}

\section{Proposed Approach}

\begin{figure*}
    \centering
    \includegraphics[width=0.9\linewidth]{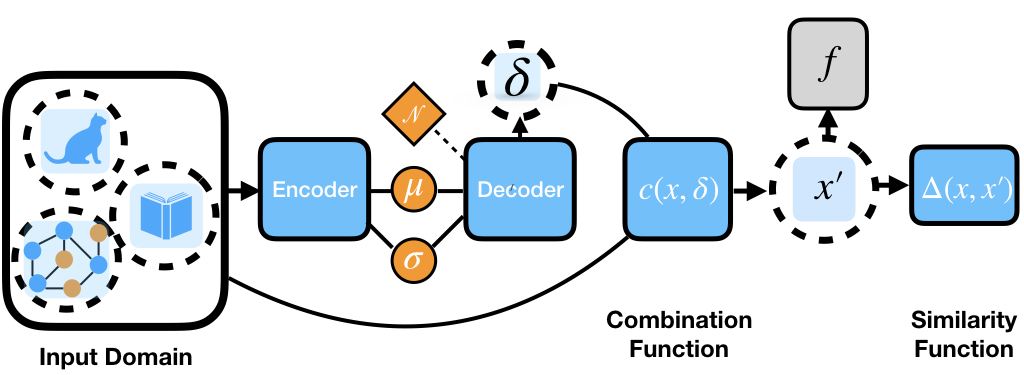}
    \caption{The main components of \namelong \ and the forward generation of an adversarial example. }
    \label{forward_gen}
    \vspace{-10pt}
\end{figure*}

To address the limitations of the constrained optimization paradigm, we propose an alternative approach based upon latent variable modeling.
The key insight in our framework is that we learn a deep generative model of our target domain, but instead of searching over the original input space $\mathcal{X}$ to generate an adversarial example, we learn to generate perturbations within a low-dimensional, continuous latent space $\mathbb{R}^d$. 
The latent codes within this space are learned by a domain specific encoder network.  

\subsection{Model overview}
We seek to define an adversarial generator network, $G$, which specifies a conditional distribution $P(x' | x)$ over adversarial examples $x'$ given an input datapoint $x$.
Without loss of generality, we define a generic adversarial generator network as a combination of four components (Figure \ref{forward_gen}):
\begin{itemize}[leftmargin=15pt, topsep=5pt, itemsep=2pt, parsep=5pt]
    \item A {\em probabilistic encoder network}, $q_\phi(z | x)$ that defines a conditional distribution over latent codes $z \in \mathbb{R}^d$ given an input $x_i \in \mathcal{X}$.  To minimize stochastic gradient variance, we assume that this encoder computes a pathwise derivative via the reparameterized Gaussian distribution \cite{kingma2014auto}.
    That is, we specify the latent conditional distribution as $z \sim \mathcal{N}(\mu_\phi(x), \sigma_\phi(x))$, where $\mu_\phi, \sigma_\phi : \mathcal{X} \rightarrow \mathbb{R}^d$ are differentiable neural networks.
    We sample from $z \sim q_\phi(z | x)$ by drawing $\epsilon \sim \mathcal{N}(0,I)$ and returning $z = \mu_\phi(x) + \sigma_\phi(x) \circ \epsilon$, where $\circ$ denotes the elementwise product. 
    \item A {\em probabilistic decoder network}, $p_\theta(\delta | z)$ that maps a sampled latent code $z \in \mathbb{R}^d$ to a perturbation $\delta$. This function is an arbitrary differentiable neural network. 
    \item A {\em combination function} $c(\delta,x)$ that takes an input $x$ and perturbation as input and outputs a perturbed example $x'$. The combination function can be as simple as adding the perturbation to the clean input---i.e., $x' = x + \delta$---or more complex such as first projecting to a candidate set.
    \item A {\em similarity function} $\Delta$ defined over $\mathcal{X}$, used to restrict the space of adversarial examples.  A common choice is any $l_p$-norm for continuous domains or an upper bound on the number of actions performed if the domain is discrete.
\end{itemize}
The basic flow of our adversarial framework is summarized in Figure \ref{forward_gen}: we use the encoder network to  sample a latent ``perturbation code'' $z \in \mathbb{R}^d$ given an input $x \in \mathcal{X}$, and given this sampled code, we then generate a perturbation $\delta = p_\theta(z)$ that is combined with our original input using the combination function to generate the adversarial sample $x' = c(\delta, x)$.

\cut{
\begin{table*}[t]
\caption{Summary of the different components to applying our framework to different input domains}
\label{tab:components}
\begin{center}
\begin{small}
\begin{sc}
\begin{tabular}{lcccccr}
\toprule
Attack Domain & Input Type & Encoder & Decoder & $\delta$& $\Delta$ & $c(x,\delta$)   \\
\midrule
Image  & Pixels & CNN & CNN & Pixels &$l_2$ & $x + \delta$  \\
Text-Emb & Word Emb& LSTM & LSTM & Word Emb &$l_2$ & $x + \delta$ \\
Text-Token & Discrete Token & LSTM & LSTM& Word Emb& $l_2$ & $ \Pi_{\mathcal{X}}(x+\delta)$\\
Graph-Direct&  Node Emb  & GCN & MLP& Node Emb &$l_2$ &$x + \delta$ \\
Graph-Influencer &  Node Emb  & GCN & MLP& Node Emb &$l_2$ &$x + b \cdot \delta$ \\
\bottomrule
\end{tabular}
\end{sc}
\end{small}
\end{center}
\vskip -0.1in
\label{attack_summary}
\end{table*}
}

This generic latent variable framework for adversarial attacks can be easily adapted to different domains (e.g., text, images, or graphs) by simply specifying domain-specific variants of the four components above, with further details provided in Section \ref{sec:domain_imp}.
A substantial benefit of this framework is that the attacker can make use of domain-specific prior information readily available in the input data through an appropriate choice of architecture. For example, if the input is an image, its inherent structure can be exploited by using the inductive bias of CNNs. Similarly, an appropriate choice for text data can be LSTMs and GNNs for graph-structured data.

\xhdr{Adversarial generalization}
Unlike the constrained optimization approach, our framework also allows us to amortize the inference process: instead of performing a specific optimization for each attacked example, our trained encoder-decoder can efficiently generate a distribution of adversarial examples on arbitrary input points, even points that were unseen during training. 
In addition, since the output of our encoder $q_\phi(z|x)$ is stochastic, we learn a {\em distribution} over adversarial examples, making it possible to efficiently sample a diverse set of attacks.
One key benefit of this fact is that even when one adversarial sample generated through \namelong\ is unsuccessful, it is computationally inexpensive to generate further samples that may be adversarial. 
In Section \ref{q3}, we exploit this phenomenon to resample new examples in cases where our first attempt fails to attack a model, and show that this strategy can significantly improve attack generalization on a test set.

\subsection{Training and loss function}
To train our model we define a hybrid objective function that is composed of a misclassification loss, $\mathcal{L}_{c}$, that penalizes the model if the generated $x'$ points are not adversarial, as well as two regularizers. 

\xhdr{Misclassification loss}
We use a max-margin misclassification loss which has been shown to be robust to various hyperparameter settings \cite{carlini2017towards}.
Using $s(x,y) \in \mathbb{R}$ to denote the (unnormalized) relative likelihood that the classifier $f$ assigns to point $x \in \mathcal{X}$ belonging to class $y \in \mathcal{Y}$, we define the classification loss as:
\begin{equation*}
    \mathcal{L}_c = \frac{1}{N}\sum_{i}^{N}\max(s(x'_i, y_i) - (s(x'_i, y'_i))_{\max_{y'_i\neq y_i}}, 0),
\end{equation*}
where $y_i$ is the correct class for the unperturbed point $x_i$.

\xhdr{Regularization}
To regularize the output perturbation, we penalize the model according to the similarity $\Delta(x, x')$ between the perturbed and unperturbed points.
We also add a KL-regularizer in the latent space between a uniform Gaussian prior, $p(z)$, and the distribution defined by $z \sim q_\phi(z | x)$, which prevents the latent samples $z$ from collapsing onto one specific latent code. 

\xhdr{Overall objective and training}
The overall objective is thus defined as follows:
\begin{align}
    \mathcal{L} = \mathcal{L}_{c} - \frac{\lambda}{N}\sum_i^N (\Delta(x_i,x'_i) + D_{KL}(q_\phi(z|x_i) \: || \: p(z)).
\label{eq:overall_obj}
\end{align}
Learning in this setting consists of first sampling a mini-batch of unperturbed samples and then running them through the adversarial generator $G$, which attempts to fool every example in the mini-batch before updating in an end-to-end fashion through stochastic gradient descent. 
Our approach is related to previous work on generative modeling with latent variable models  \cite{kingma2014auto}. However, our objective function is an adversarial loss and not the usual cross-entropy objective. Additionally, the latent variable is modeled as an additive perturbation distribution, as opposed to merely a latent conditional variable for the generator.
Moreoever, the reconstruction error in Equation \eqref{eq:overall_obj} is given by an arbitrary similarity function $\Delta$, with a hyperparameter $\lambda$ that trades-off the adversarial misclassification objective from the magnitude of the perturbation by maximizing the similarity.

\subsection{Implementations in different input domains}\label{sec:domain_imp}

To deploy the \namelong\ framework on a particular input domain, one simply needs to specify domain-specific implementations of the four key components: an encoder network, decoder network, combination function, and a similarity function. 
%

\xhdr{Attacks on image classification}
Following standard practice in the image domain, we define the encoder $q_\phi(z | x)$ to be a convolutional neural network (CNN) and the decoder $p_\theta(\delta|z)$ to be a deconvolutional network. 
The combination function is defined to be simple addition $c(\delta, x) = x + \delta$ and we define $\Delta$ to be the $l_2$ distance.
Thus, in this setting our adversarial generator network $G$ uses a generative model to output a small perturbation in continuous pixel space. 

\xhdr{Attacks on text classification}
In the text setting, we define the input $x=(x_1, ..., x_T)$ to be a sequence of $T$ words from a fixed vocabulary $\mathcal{V}$, and we seek to generate $x'$ as an equal-length perturbed sequence of words.
Since we are focusing on attacking recurrent neural networks (RNNs), we also assume that every word $w \in \mathcal{V}$ is associated with an embedding vector $E(w) \in \mathbb{R}^d$, via a known embedding function $E : \mathcal{V} \rightarrow \mathbb{R}^d$.
We use an LSTM encoder $q_\phi(z | x)$ to map the sequence $(E(x_1), ..., E(x_T))$ of input word embeddings to a latent vector $z$.
The decoder $p_\theta(\delta | z)$ is then defined as an LSTM that maps the latent code $z$ to a sequence of $T$ perturbation vectors $\delta = (\delta_1, ..., \delta_T)$, with $\delta_i \in \mathbb{R}^d$.
To generate a valid sequence of words, we combine each perturbation $\delta_i$ with the corresponding input word embedding $E(x_i)$ and employ a projection function $\Pi_{\mathcal{X}} : \mathbb{R}^d \rightarrow \mathcal{V}$ to map each perturbed embedding to valid vocabulary word via a nearest-neighbor lookup in the embedding space. 

Note, however, that the discrete nature of the projection operator $\Pi_\mathcal{X}$ represents a challenge for end-to-end training.
One option is to simply define the combination function as $c(\delta_i, x_i)=\Pi_{\mathcal{X}}(E(x_i) + \delta_i)$ and omit the discrete projection during training (similar to \cite{gong2018adversarial}), but---as we will show in the experiments (Section \ref{q1})---this leads to sub-optimal attacks. 

To address this issue, we develop a differentiable nearest neighbour attention mechanism (inspired by \cite{plotz2018neural}), where for each word $x_i$ in the input sequence $x$, we define the combination function $c(x_i, \delta_i)$ with a tunable temperature parameter $\tau$ as:
\begin{equation}\label{eq:diffnn}
    x'_i = c(\delta_i, x_i) = \Pi_{\mathcal{X}}\left(\sum_{w \in \mathcal{V}} \frac{\exp(\alpha_w/\tau)E(w)}{\sum_{w' \in \mathcal{V}} \exp(\alpha_{w'} / \tau)}\right)
\end{equation}
where $\alpha_w$ denotes the angular distance between the perturbed embedding $E(x_i) + \delta_i$ and the embedding $E(w)$ of vocabulary word $w \in \mathcal{V}$. 
As with images, we use the $l_2$ norm on the embeddings as the similarity function. 

\xhdr{Attacks on node classification}
The final domain we consider is the setting of semi-supervised node classification using graph neural networks (GNNs) \cite{kipf2016semi}.
We consider the challenging attack setting where the attack model can only make changes to node attributes and not the graph structure itself \cite{zugner2018adversarial}---employing a graph convolution network (GCN) \cite{kipf2016semi} as the encoder network and a multi-layer perceptron (MLP) as the decoder.
Note, however, that adversarial attacks on graphs present unique complications compared to texts and images in that the training data is non-i.i.d., with the training points being a sub-sample of the nodes in a large graph.
Thus, following Zugner et al. \cite{zugner2018adversarial}, we consider both {\em direct attacks}, which modify the features of the target nodes themselves as well as {\em influencer attacks}, which can only modify features in the neighborhood of a target node but not the node itself.
Consequently, we define two sets of disjoint nodes: the attacker set $\mathcal{A}$, and the target set $\mathcal{T}$. 
For direct attacks $\mathcal{A} = \mathcal{T}$ and in this case the combination function is simply $c(\delta, x) = x + \delta$. For influencer attacks, only the embeddings of the $\mathcal{A}$ are modified and thus we use a binary mask in our combination function, i.e., $c(\delta, x) = x + b \cdot \delta$, where $b \in [0,1]^N$.
We use the $l_2$ norm as the similarity function.

\section{Experiments}\label{sec:experiments}
We investigate the application of our generative framework to produce adversarial examples against classification tasks where the input domain is one of natural images, textual data and graph-structured data. Through our experiments we seek to answer the following questions: 
\begin{enumerate}[label=({\bf Q\arabic*}), leftmargin=28pt, topsep=2pt, itemsep=2pt, parsep=2pt]
\item
 \textbf{Domain Agnostic.} Can we use the same attack strategy across multiple domains?
\item
\textbf{Attack Generalization.} Is it possible to adversarially generalize to unseen examples?
\item \textbf{Diversity of Attacks.} Can we generate diverse sets of adversarial examples?
\end{enumerate}
As points of comparison throughout our experiments,  we consider constrained optimization-based baselines specific to the different domains we examine (and which use comparable setups).
We also experiment with a simplified version of \namelong\ that uses a deterministic autoencoder, rather than a variational approach (denoted \aename). 
Code to reproduce our experiments is included with the submission and will be made public.

\begin{figure*}
\centering
    \includegraphics[width=0.32\linewidth]{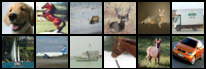}
    \includegraphics[width=0.32\linewidth]{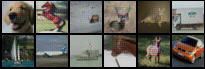}
    \includegraphics[width=0.32\linewidth]{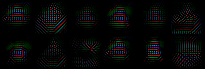}
    \caption{\textbf{Left} Unperturbed images from CIFAR10. \textbf{Middle} Attacked images with $\lambda=0.1$. \textbf{Right}: Perturbation magnified by a factor of 10 for sake of visualization.}
    \label{fig:cifar_01}
\end{figure*}

\subsection{Datasets and Target Models}
We attack popular neural network classification models across image, text, and graph classification using standard benchmark datasets.

\xhdr{Image domain}
We use the CIFAR10 dataset for image classification (with standard train/test splits), and attack a CNN classification model based on the VGG16 architecture.
The target CNN model is trained for $100$ epochs with the Adam optimizer with learning rate fixed at $1e-4$.

\xhdr{Text domain}
We use the IMDB dataset for sentiment classification (with standard train/test splits) \cite{maas-EtAl:2011:ACL-HLT2011}. 
The target classifier is a single-layer LSTM model, initialized with pretrained GloVe embeddings and  trained for $10$ epochs with the Adam optimizer at default settings.
Note that we focus on {\em word-level} attacks and omit a detailed comparison to character-level heuristic approaches (e.g., \cite{li2018textbugger,ebrahimi2017hotflip}).

\xhdr{Graph domain}
We consider two standard node classification datasets, Cora and CiteSeer \cite{lu2003link,bhattacharya2007collective}. 
We split the dataset into a labeled $20\%$, of which $10\%$ is used for training and the remaining is used for validation. The remaining nodes are unlabeled nodes and are used for testing purposes. We attack a single-layer graph convolutional network (GCN) model \cite{kipf2016semi} that is trained for $100$ epochs with the Adam optimizer and a learning rate of $10^{-2}$.
Following previous work, we consider both {\em direct} and {\em influencer} attacks (as discussed in Section \ref{sec:domain_imp}); in the influencer setting we attack a node by changing the features of a random sample of $5$ of its neighbors.\footnote{If there are fewer than $5$ neighbors we randomly sample nodes from $\mathcal{A}$ until we have a total of $5$}

\subsection{Results}
We now address the core experimental questions (\textbf{Q1}-\textbf{Q3}), highlighting the peformance of \namelong\ across the three distinct domains. 

\xhdr{Q1: Domain Agnostic}
\label{q1}
Our first set of experiments focus on demonstrating that \namelong \ can achieve strong performance across three distinct input domains.
Overall, we find that \namelong\ performs competitively in the image setting (Table \ref{tab:image}) and achieves very strong results in the text (Table \ref{tab:text}) and graph (Table \ref{tab:graph}) domains.
However, we emphasize that these results are meant to demonstrate the flexibility of \namelong\ across distinct domains; an exhaustive comparison within each domain---especially for images---would necessitate a greater variety of benchmarks and evaluation settings.

For the image domain our baselines are the $l_2$-variants of the PGD \cite{madry2017towards} and Carlini-Wagner \cite{carlini2017towards} attacks.
We find that \namelong \ performs competitively with the baselines with a small drop of $8\%$ in performance for the \namelong\ model and a smaller drop of $3\%$ for the \aename\ model (Table \ref{tab:image}).
Figure \ref{fig:cifar_01} provides an example of the perturbations generated by our approach. 

For the text domain, we report the results with and without the differentiable nearest neighbor mechanism (Equation \ref{eq:diffnn}), with the latter also applied to the \aename\ baseline for fair comparison. 
We compare against results reported in Gong et al.\@ \cite{gong2018adversarial} for their FGSM baseline and DeepFool model.
We find that \namelong\ outperforms both of these baselines, with the differentiable nearest-neighbour mechanism being crucial to prevent overfitting (Table \ref{tab:text}\cut{see Table \ref{tab:text_example} for an example attack, with further details in the Appendix}).
 Figure \ref{fig:lambda} plots the tradeoff in adversarial success between the perturbation regularization and percentage of tokens changed. As expected, more regularization reduces the number of words perturbed, at the expense of lower performance.
Note that we do not compare against recently reported results on Alzantot et al.'s genetic attack algorithm \cite{alzantot2018generating}, since it has not been scaled beyond a small subset of the IMDB data and requires significantly truncating the review text.

Finally, in the graph domain, we find that our \aename\ baseline achieves a new state-of-the-art for {\em direct attacks} (when modifying node features only), outperforming the constrained optimization NetAttack approach \cite{zugner2018adversarial} and achieving a perfect success rate on the training set (Table \ref{tab:graph}).
Both \aename\ and \namelong\ also significantly outperform NetAttack in the {\em influencer attack} setting, with the latter seeing an absolute improvement of $29\%$ and $16\%$ on Cora and CiteSeer, respectively. 
Note that we do not compare against the graph attack framework of Dai et al.\@ \cite{dai2018adversarial}, since that work modifies the adjacency matrix of the graph, whereas we modify node features only.

\xhdr{Q2: Attack Generalization}
\label{q2}
We demonstrate the ability of \namelong\ to a attack a test set of previously {\em unseen} instances, with results summarized under the Test columns in Tables 2-4. Since constrained optimization-based attack strategies cannot generalize to a test set of examples, we rely on our \aename\ approach as a strong baseline. In all three domains \namelong\ is able to generalize effectively, outperforming \aename\ in the image setting and in the text setting. 
For graph attacks we observe that \namelong\ has marginally better generalization ability for influencer attacks while \aename\ performs better by a similar margin in the easier direct attack setting.

\xhdr{Q3: Diversity of Attacks}
\label{q3}
\namelong\ exhibits comparable (or marginally worse) performance compared to our deterministic autoencoder (\aename) approach in terms of raw success rates.
However, a key benefit of the full \namelong\ framework is that it has a stochastic latent state, which allows for resampling of latent codes to produce new adversarial examples. 
We empirically verify this phenomena by resampling failed test set examples up to a maximum of $100$ resamples. 
As can be seen from Figure \ref{fig:resampling}, \namelong \ can produce adversarial samples for clean inputs that were originally classified correctly, significantly boosting generalization performance. Specifically, we observed that $71\%$ of correctly classified test samples could be successfully attacked for CIFAR10 after $100$ resamples. Similarly, for IMDB there is an average absolute improvement of $60\%$. For the graph setting we resample test set instances for direct attacks and failed training set instances for influencer attacks but still without any further optimization. We achieve significant improvements of $36\%$ and $61\%$ for Cora and CiteSeer direct attacks, respectively, and $47\%$ and $8\%$ for Cora and CiteSeer influencer attacks, respectively.
\cut{
\begin{table*}[t]
        \begin{minipage}{0.99\linewidth}
                \begin{small}
                  \caption{Attack success rate of \namelong\  and \aename\ on CIFAR-10 for Training and Test splits. We used $\lambda=0.1$ and the test set results are averaged over 10 runs.}
                  \label{tab:image}
            \begin{tabular}{ccccc}
            \toprule
                   & \nameshort\ & \aename\ & PGD \cite{madry2017towards} & Carlini-Wagner \cite{carlini2017towards} \\
            \midrule
                   Train & 92\% & 97\% & \textbf{100\%} & \textbf{100\%}\\
                   Test & \textbf{81\%} & 67\% & - & - \\
            \bottomrule
            \end{tabular}
          
            \end{small}
        \end{minipage}
    \end{table*}
}
\begin{table}[t]
        \begin{minipage}{0.49\textwidth}
                \centering
                \begin{small}
                  \caption{Attack success rate of \namelong\  and \aename\ on CIFAR-10 for Training and Test splits. We used $\lambda=0.1$ and the test set results are averaged over 10 runs.}
                  \label{tab:image}
            \begin{tabular}{lcc}
            \toprule
                    & Cifar-Train & Cifar-Test \\
            \midrule
                    \nameshort\ & 92\% & \textbf{81\%}\\
                    \aename\ & 97\% & 67\% \\
                    PGD \cite{madry2017towards} & \textbf{100\%} & -\\
                    Carlini-Wagner \cite{carlini2017towards} & \textbf{100\%}& -\\
                    
            \bottomrule
            \end{tabular}
          
            \end{small}
        \end{minipage}
\end{table}

 \begin{table}[t]
        \begin{minipage}{0.49\textwidth}
        \centering 
         \begin{small}
         \caption{Attack success rate of \namelong\ and \aename\ on IMDB for Train and Test splits.  We used $\lambda=0.05$ and and cap the number words changed to $15\%$ of the sample length.} 
         \label{tab:text}
            \begin{tabular}{lcc}
                \toprule
                & IMDB-Train & IMDB-Test \\
                \midrule
                \aename\ &  6\% & 18\% \\
                \aename-Diff & 44\% & 45\% \\
                \nameshort & 9\% & 20\%\\ 
                \nameshort-Diff &  \textbf{71\%} & \textbf{61\%}\\
                FGSM \cite{gong2018adversarial} & 32\% & -\\
                DF  \cite{gong2018adversarial} & 36\% & -\\
            \bottomrule
            \end{tabular}
        \end{small}
        \vspace{-10pt}
        \end{minipage}
\end{table} 

\begin{table}[t]
\begin{minipage}{0.48\textwidth}
 \begin{scriptsize}
            \caption{Attack success rate of \namelong\ and \aename\ on Graph Data. For Influencer attacks we only consider the training set. We used $\lambda=0.01$ and results are averaged over 5 runs.}
            \label{tab:graph}
            \begin{tabular}{lccccr}
                \toprule
                  &  Cora Train &  Cora Test  &  CiteSeer Train & CiteSeer  Test\\
                \midrule
                \namelong-Direct  & 88 \%& 91 \% & 81\% & 82\% \\
                \aename-Direct & \textbf{100\%} & \textbf{93\%} & \textbf{100\%} & \textbf{92\%} \\
                Zugner-Direct & 99\% & - & 99\%&-\\
                \midrule
                \namelong-Inf & \textbf{62\%} &  - & \textbf{54 \%} & - \\
                \aename-Inf & 60\%  & - & 52\% & -\\
                Zugner-Inf \cite{zugner2018adversarial} & 33 \% & - & 38\%&-\\
            \bottomrule
            \end{tabular}
            \end{scriptsize}
    \end{minipage}
\end{table}

\begin{figure}[t!]
    \begin{minipage}{0.48\textwidth}
     \centering
    \includegraphics[width=1\linewidth]{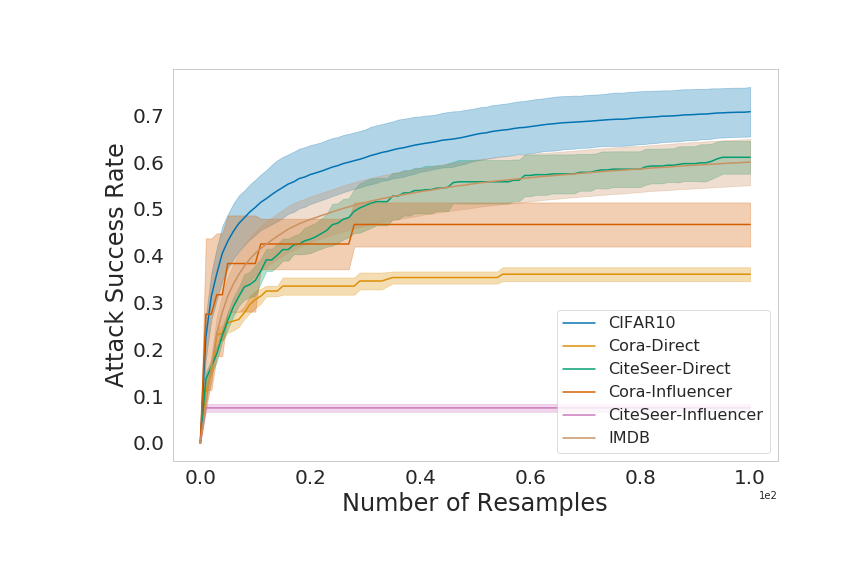}
    \vspace{-10pt}
    \caption{Attack Success Rate when resampling only failed adversarial examples in the test set.}
     \label{fig:resampling}
    \end{minipage}\hfill
     \begin{minipage}{0.48\textwidth}
      \centering
    \includegraphics[width=0.85\linewidth]{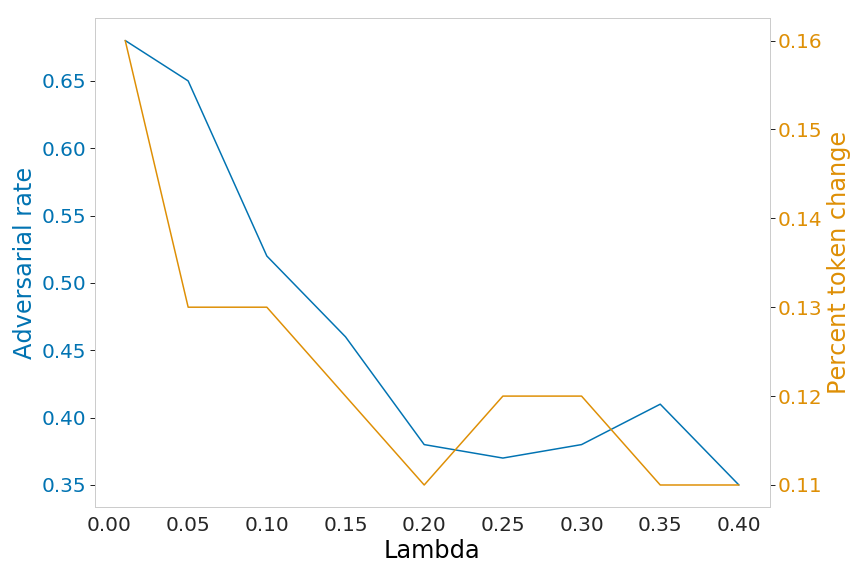}
    \caption{Tradeoff in attack success rate and word error rate with respect to $\lambda$.}
  \label{fig:lambda}
    \end{minipage}\hfill
    \vspace{-1mm}
\end{figure}

\section{Further Related Work}
Here, we provide a brief discussion of related work that we have not previously discussed (e.g, in Sections \ref{sec:intro} and \ref{sec:experiments}), and that is highly relevant to our proposed framework.
For more comprehensive overviews of recent advancements in adversarial attacks, we direct the interested reader to survey papers for adversarial attacks in the image \cite{akhtar2018threat}, text \cite{text_survey1}, and graph domains \cite{sun2018adversarial}.

\xhdr{Adversarial attacks using parametric models}
Similar to our approach, there is relevant previous work on generating adversarial images using parametric models, such as the Adversarial Transformation Networks (ATN) framework and its variants \cite{baluja2017adversarial,xiao2018generating}.
However, unlike our approach, they are specific to the image domain and do not define a generative distribution over adversarial examples.

\xhdr{Adversarial attacks in latent space}
There is a burgeoning literature of attacks that are crafted in a latent space, similar to the framework proposed here.
For instance, Zhao et al.\@ \cite{zhao2017generating} propose a framework to generate ``natural'' adversarial examples that lie on the data manifold.
However, unlike our work, their ``natural adversary'' requires a search procedure to generate an adversarial example and does not facilitate an efficient generation of diverse attacks like \namelong. 
Concurrent to our work, Li et al.\@ \cite{li2019nattack} introduced $\mathcal{N}\textrm{Attack}$ to produce distributions of adversarial examples for  blackbox image-based attacks.
While similar in spirit to our framework and relatively efficient, $\mathcal{N}\textrm{Attack}$ is specific to the image domain and operates within a constrained optimization paradigm---requiring additional rounds of optimization to generate adversarial examples on unseen data.
\section{Discussion and Conclusion}
We present \namelong, a unified framework for constructing domain agnostic adversarial attacks. We deploy our strategy on three domains---images, text and graphs---and successfully show that our trained model is capable of generating adversarial examples on unseen test examples in a new form of attack generalization. Furthermore, we show that by learning perturbations as a latent variable, our framework is capable of generating diverse sets of attacks for any given input, allowing the attacker to cheaply resample new attacks in the event that an initial attacks fails. 

In this work, we have considered attacks where the latent perturbation is modeled as a reparametrized Gaussian. In the future, we would like to extent \namelong\ to other distributions using various generative modelling methods, such as normalizing flows. 
Extending \namelong\ to the blackbox setting (e.g., using relaxed gradient estimators) is another fruitful direction for future work.
Finally, while this work demonstrates the significant promise of a latent variable approach to adversarial attacks, further empirical studies, e.g.  using various adversarial defense strategies, are necessary to more completely characterize the pros and cons of this approach compared to the constrained optimization paradigm.

\clearpage
\bibliography{bibliography.bib}

\begin{thebibliography}{10}

\bibitem{akhtar2018threat}
Naveed Akhtar and Ajmal Mian.
\newblock Threat of adversarial attacks on deep learning in computer vision: A
  survey.
\newblock {\em IEEE Access}, 6:14410--14430, 2018.

\bibitem{alzantot2018generating}
Moustafa Alzantot, Yash Sharma, Ahmed Elgohary, Bo-Jhang Ho, Mani Srivastava,
  and Kai-Wei Chang.
\newblock Generating natural language adversarial examples.
\newblock {\em arXiv preprint arXiv:1804.07998}, 2018.

\bibitem{baluja2017adversarial}
Shumeet Baluja and Ian Fischer.
\newblock Adversarial transformation networks: Learning to generate adversarial
  examples.
\newblock {\em arXiv preprint arXiv:1703.09387}, 2017.

\bibitem{bhattacharya2007collective}
Indrajit Bhattacharya and Lise Getoor.
\newblock Collective entity resolution in relational data.
\newblock {\em ACM Transactions on Knowledge Discovery from Data (TKDD)},
  1(1):5, 2007.

\bibitem{carlini2017towards}
Nicholas Carlini and David Wagner.
\newblock Towards evaluating the robustness of neural networks.
\newblock In {\em Security and Privacy (SP), 2017 IEEE Symposium on}, pages
  39--57. IEEE, 2017.

\bibitem{dai2018adversarial}
Hanjun Dai, Hui Li, Tian Tian, Xin Huang, Lin Wang, Jun Zhu, and Le~Song.
\newblock Adversarial attack on graph structured data.
\newblock {\em arXiv preprint arXiv:1806.02371}, 2018.

\bibitem{ebrahimi2017hotflip}
Javid Ebrahimi, Anyi Rao, Daniel Lowd, and Dejing Dou.
\newblock Hotflip: White-box adversarial examples for text classification.
\newblock {\em arXiv preprint arXiv:1712.06751}, 2017.

\bibitem{gao2018black}
Ji~Gao, Jack Lanchantin, Mary~Lou Soffa, and Yanjun Qi.
\newblock Black-box generation of adversarial text sequences to evade deep
  learning classifiers.
\newblock In {\em 2018 IEEE Security and Privacy Workshops (SPW)}, pages
  50--56. IEEE, 2018.

\bibitem{gong2018adversarial}
Zhitao Gong, Wenlu Wang, Bo~Li, Dawn Song, and Wei-Shinn Ku.
\newblock Adversarial texts with gradient methods.
\newblock {\em arXiv preprint arXiv:1801.07175}, 2018.

\bibitem{goodfellow2014explaining}
Ian~J Goodfellow, Jonathon Shlens, and Christian Szegedy.
\newblock Explaining and harnessing adversarial examples.
\newblock {\em arXiv preprint arXiv:1412.6572}, 2014.

\bibitem{kingma2014auto}
Diederik~P Kingma and Max Welling.
\newblock Auto-encoding variational bayes.
\newblock In {\em International Conference on Learning Representations (ICLR)},
  2014.

\bibitem{kipf2016semi}
Thomas~N Kipf and Max Welling.
\newblock Semi-supervised classification with graph convolutional networks.
\newblock {\em arXiv preprint arXiv:1609.02907}, 2016.

\bibitem{li2018textbugger}
Jinfeng Li, Shouling Ji, Tianyu Du, Bo~Li, and Ting Wang.
\newblock Textbugger: Generating adversarial text against real-world
  applications.
\newblock {\em arXiv preprint arXiv:1812.05271}, 2018.

\bibitem{li2019nattack}
Yandong Li, Lijun Li, Liqiang Wang, Tong Zhang, and Boqing Gong.
\newblock Nattack: Learning the distributions of adversarial examples for an
  improved black-box attack on deep neural networks.
\newblock {\em arXiv preprint arXiv:1905.00441}, 2019.

\bibitem{lu2003link}
Qing Lu and Lise Getoor.
\newblock Link-based classification.
\newblock In {\em Proceedings of the 20th International Conference on Machine
  Learning (ICML-03)}, pages 496--503, 2003.

\bibitem{maas-EtAl:2011:ACL-HLT2011}
Andrew~L. Maas, Raymond~E. Daly, Peter~T. Pham, Dan Huang, Andrew~Y. Ng, and
  Christopher Potts.
\newblock Learning word vectors for sentiment analysis.
\newblock In {\em Proceedings of the 49th Annual Meeting of the Association for
  Computational Linguistics: Human Language Technologies}, pages 142--150,
  Portland, Oregon, USA, June 2011.

\bibitem{madry2017towards}
Aleksander Madry, Aleksandar Makelov, Ludwig Schmidt, Dimitris Tsipras, and
  Adrian Vladu.
\newblock Towards deep learning models resistant to adversarial attacks.
\newblock {\em arXiv preprint arXiv:1706.06083}, 2017.

\bibitem{moosavi2016deepfool}
Seyed-Mohsen Moosavi-Dezfooli, Alhussein Fawzi, and Pascal Frossard.
\newblock Deepfool: a simple and accurate method to fool deep neural networks.
\newblock In {\em Proceedings of the IEEE Conference on Computer Vision and
  Pattern Recognition}, pages 2574--2582, 2016.

\bibitem{papernot2016transferability}
Nicolas Papernot, Patrick McDaniel, and Ian Goodfellow.
\newblock Transferability in machine learning: from phenomena to black-box
  attacks using adversarial samples.
\newblock {\em arXiv preprint arXiv:1605.07277}, 2016.

\bibitem{papernot2017practical}
Nicolas Papernot, Patrick McDaniel, Ian Goodfellow, Somesh Jha, Z~Berkay Celik,
  and Ananthram Swami.
\newblock Practical black-box attacks against machine learning.
\newblock In {\em Proceedings of the 2017 ACM on Asia Conference on Computer
  and Communications Security}, pages 506--519. ACM, 2017.

\bibitem{papernot2016limitations}
Nicolas Papernot, Patrick McDaniel, Somesh Jha, Matt Fredrikson, Z~Berkay
  Celik, and Ananthram Swami.
\newblock The limitations of deep learning in adversarial settings.
\newblock In {\em Security and Privacy (EuroS\&P), 2016 IEEE European Symposium
  on}, pages 372--387. IEEE, 2016.

\bibitem{plotz2018neural}
Tobias Pl\"{o}tz and Stefan Roth.
\newblock Neural nearest neighbors networks.
\newblock In {\em Advances in Neural Information Processing Systems 31}, pages
  1087--1098, 2018.

\bibitem{sun2018adversarial}
Lichao Sun, Ji~Wang, Philip~S Yu, and Bo~Li.
\newblock Adversarial attack and defense on graph data: A survey.
\newblock {\em arXiv preprint arXiv:1812.10528}, 2018.

\bibitem{szegedy2013intriguing}
Christian Szegedy, Wojciech Zaremba, Ilya Sutskever, Joan Bruna, Dumitru Erhan,
  Ian Goodfellow, and Rob Fergus.
\newblock Intriguing properties of neural networks.
\newblock {\em arXiv preprint arXiv:1312.6199}, 2013.

\bibitem{text_survey1}
Wenqi Wang, Benxiao Tang, Run Wang, Lina Wang, and Aoshuang Ye.
\newblock A survey on adversarial attacks and defenses in text.
\newblock {\em CoRR}, abs/1902.07285, 2019.

\bibitem{xiao2018generating}
Chaowei Xiao, Bo~Li, Jun-Yan Zhu, Warren He, Mingyan Liu, and Dawn Song.
\newblock Generating adversarial examples with adversarial networks.
\newblock {\em arXiv preprint arXiv:1801.02610}, 2018.

\bibitem{zhao2017generating}
Zhengli Zhao, Dheeru Dua, and Sameer Singh.
\newblock Generating natural adversarial examples.
\newblock {\em arXiv preprint arXiv:1710.11342}, 2017.

\bibitem{zugner2018adversarial}
Daniel Z{\"u}gner, Amir Akbarnejad, and Stephan G{\"u}nnemann.
\newblock Adversarial attacks on neural networks for graph data.
\newblock In {\em Proceedings of the 24th ACM SIGKDD International Conference
  on Knowledge Discovery \& Data Mining}, pages 2847--2856. ACM, 2018.

\end{thebibliography}
\bibliographystyle{plain}
\end{document}